# Coping with the Limitations of Rational Inference in the Framework of Possibility Theory

**Salem Benferhat, Didier Dubois and Henri Prade**
Institut de Recherche en Informatique de Toulouse (IRIT) – CNRS
Université Paul Sabatier, 118 route de Narbonne
31062 Toulouse Cedex, France
Email: {benferhat, dubois, prade}@irit.fr

### Abstract
Possibility theory offers a framework where both Lehmann's "preferential inference" and the more productive (but less cautious) "rational closure inference" can be represented. However, there are situations where the second inference does not provide expected results either because it cannot produce them, or even provide counter-intuitive conclusions. This state of facts is not due to the principle of selecting a unique ordering of interpretations (which can be encoded by one possibility distribution), but rather to the absence of constraints expressing pieces of knowledge we have implicitly in mind. It is advocated in this paper that constraints induced by independence information can help finding the right ordering of interpretations. In particular, independence constraints can be systematically assumed with respect to formulas composed of literals which do not appear in the conditional knowledge base, or for default rules with respect to situations which are "normal" according to the other default rules in the base. The notion of independence which is used can be easily expressed in the qualitative setting of possibility theory. Moreover, when a counter-intuitive plausible conclusion of a set of defaults, is in its rational closure, but not in its preferential closure, it is always possible to repair the set of defaults so as to produce the desired conclusion.

## 1 INTRODUCTION

It seems that there is now a general agreement about the foundations of exception-tolerant reasoning which have been proposed by Lehmann and his colleagues (Kraus et al., 1990; Lehmann and Magidor, 1992) and by Gärdenfors and Makinson (1994) who provide basic systems of postulates for nonmonotonic consequence relations. From these proposals two systems are particularly emerging: on the one hand the System P (P for preferential) which offers a basic core for commonsense reasoning but provides a very cautious inference machinery. On the other hand, the rational closure inference proposed by Lehmann and Magidor (1992) which is based on System P augmented with a rational monotony postulate, provides more adventurous conclusions and remains more controversial. Indeed, although the rational closure inference often gives expected conclusions, examples have been pointed out showing that this inference may still be too cautious in some situations, and another example is recalled in this paper showing that it may as well lead to counter-intuitive conclusions. In previous papers (Benferhat et al., 1992; Dubois and Prade, 1995), the authors have shown that a simple semantics can be provided for the two types of nonmonotonic inferences (preferential and rational closure types) in the framework of possibility theory. Using the semantical framework of possibility theory, this paper explores the limitations of rational closure and shows that these limitations do not concern the principle of rational monotony as such but rather to the fact that some implicit knowledge, like independence, is not expressed. Independence information can be also represented in the possibilistic framework under the form of additional constraints on the plausibility ordering existing between the possible states of the world. Introducing independence information provides a natural way for getting desirable conclusions which cannot be obtained by the rational closure inference without expressing the constraints indu-ced by independence. Related points of view for improving plausible inference have been recently expressed by Delgrande and Pelletier (1994), Tan and Pearl (1995).

We first briefly recall the representation of preferential and rational closure inferences in the possibility theory setting before introducing and discussing independence in this framework.

## 2 DEFAULT AND STRICT RULES

### 2.1 Possibility Theory and Defaults

By a conditional assertion (we also call it a default rule) we mean a general rule of the form "generally, if $\alpha$ then $\beta$" having possibly some exceptions. These rules are denoted by "$\alpha \rightarrow \beta$" where $\rightarrow$ is a *non-classical* arrow relating two classical formulas. A *default base* is a set $\Delta = \{\alpha_i \rightarrow \beta_i, i=1,...,n\}$ of default rules. The material implication is denoted by $\Rightarrow$, and will be used to encode strict rules (without exceptions) of the form "if $\alpha_i$ is observed, then *certainly* $\beta_i$ is true". Let $W = \{\alpha_i \Rightarrow \beta_i / i=1,m\}$ be a set of strict rules.

In (Benferhat et al., 1992), it has been proposed to view each conditional assertion $\alpha \rightarrow \beta$ as a constraint expressing that the situation where $\alpha$ and $\beta$ is true has a greater plausibility than the one where $\alpha$ and $\neg\beta$ is true. In other words, in the context where $\alpha$ is true, $\beta$ is more possible or plausible than $\neg\beta$, i.e., the exceptional situation $\alpha\wedge\neg\beta$ is strictly less possible than the normal state of affairs which is $\alpha\wedge\beta$. This is expressed by

$$\Pi(\alpha\wedge\beta) > \Pi(\alpha\wedge\neg\beta)$$

where $\Pi$ is a possibility measure (Zadeh, 1978), i.e., is such that, for all propositions $\phi$, $\psi$, $\Pi(\phi\vee\psi)= \max(\Pi(\phi), \Pi(\psi))$. Moreover by convention the range of $\Pi$ is $[0,1]$ and $\Pi(\bot)=0$, $\Pi(\top)=1$. To see that the max-decomposability of $\Pi$ is natural in this context, note that if $\beta$ is the normal course of things in context $\alpha$, we should have $\Pi(\beta|\alpha)=1>\Pi(\neg\beta|\alpha)$ if and only if $\Pi(\alpha\wedge\beta)>\Pi(\alpha\wedge\neg\beta)$ where the conditional $\Pi(\beta|\alpha)$ obeys an equation of the form $\Pi(\alpha\wedge\beta)=f(\Pi(\beta|\alpha),\Pi(\alpha))$. Any choice for f such that $f(1,a)=a$ implies $\Pi(\alpha)=\Pi(\alpha\wedge\beta)$ when $\Pi(\alpha\wedge\beta) > \Pi(\alpha\wedge\neg\beta)$, which shows that if $\phi\wedge\psi=\bot$, then $\Pi(\phi\vee\psi)=\max(\Pi(\phi),\Pi(\psi))$, changing $\alpha\wedge\beta$ into $\phi$ and $\alpha\wedge\neg\beta$ into $\psi$. It can be proved that it is still equivalent to the unrestricted max-decomposability property dropping the condition $\phi\wedge\psi=\bot$.

Given a finite propositional language $\mathcal{L}$, a possibility measure $\Pi$ can be conveniently described by a so-called possibility distribution $\pi$ which associates to each interpretation $\omega$ of $\mathcal{L}$ its degree of possibility $\pi(\omega)$. Then $\Pi(\phi) = \max\{\pi(\omega) \mid \omega\models\phi\}$. Thus $\pi$ rank-orders the interpretations according to their plausibility to represent the real state of the world. $[0,1]$ is viewed here as a *purely ordinal* scale. A conditional possibility measure is defined as the greatest (least specific) possibility measure such that $\Pi(\alpha\wedge\beta) = \min(\Pi(\beta|\alpha), \Pi(\alpha))$, i.e., for $\alpha\neq\bot$

$$\Pi(\beta|\alpha) = 1 \text{ if } \Pi(\alpha) = \Pi(\alpha\wedge\beta)$$
$$= \Pi(\alpha\wedge\beta) \text{ if } \Pi(\alpha) = \Pi(\alpha\wedge\neg\beta) > \Pi(\alpha\wedge\beta),$$

keeping in mind that $\Pi(\alpha) = \max(\Pi(\alpha\wedge\beta), \Pi(\alpha\wedge\neg\beta))$. A necessity measure is associated to a possibility measure by duality, namely

$$N(\beta|\alpha) = 1-\Pi(\neg\beta|\alpha).$$

Here $1-(\cdot)$ is just a way of encoding an order-reversing operation. In an ordinal scale this operation just amounts to reversing the scale. Thus a conditional constraint can be equivalently written $\Pi(\alpha\wedge\beta) > \Pi(\alpha\wedge\neg\beta) \Leftrightarrow N(\beta|\alpha) > 0$.

Besides, strict rules of the form "all $\alpha$ are $\beta$" are modelled here by the constraint $\Pi(\alpha\wedge\neg\beta)=0$ (Benferhat, 1994), i.e., any situation where $\alpha\wedge\neg\beta$ is true is impossible, and hence will be ignored in the deduction process as we shall see.

## 2.2 Universal Consequence Relation

A set of beliefs $(\Delta=\{\alpha_i\rightarrow\beta_i, i=1,n\}, W=\{\alpha_j\Rightarrow\beta_j, j=1,m\})$ with consistent conditions (i.e., $\forall i, \alpha_i\neq\bot$) can be viewed as a family of constraints $\tilde{C}(\Delta,W)$ restricting a family $\Pi(\Delta,W)$ of possibility distributions *compatible* with $(\Delta,W)$:

**Definition 1.** *A possibility distribution $\pi$ associated with a possibility measure $\Pi$ is compatible with $(\Delta,W)$ iff we have*
*i) for each strict rule $\alpha_i\Rightarrow\beta_i$ of W: $\Pi(\alpha_i \wedge \neg\beta_i) = 0$,*
*ii) for each default rule $\alpha_i\rightarrow\beta_i$ of $\Delta$: $\Pi(\alpha_i\wedge\beta_i)>\Pi(\alpha_i\wedge\neg\beta_i)$.*

Note that $\Pi(\Delta,W)$ can be empty, and in this case our beliefs $(\Delta,W)$ are said to be potentially inconsistent. A typical example of a potentially inconsistent belief base is $\Delta=\{\alpha\rightarrow\beta, \alpha\rightarrow\neg\beta\}$. However, in the general case, there are several possibility distributions which are compatible with $(\Delta,W)$. The question is then how to define which conditionals $\alpha\rightarrow\beta$ are entailed from our beliefs $(\Delta,W)$. A first way to do it considers all the possibility distributions of $\Pi(\Delta,W)$, namely:

**Definition 2.** *A conditional assertion $\alpha\rightarrow\beta$ is said to be a universal possibilistic consequence of $(\Delta,W)$, denoted by $(\Delta,W) \models_{\forall\Pi} \alpha\rightarrow\beta$, if and only if $\beta$ is a possibilistic consequence of $\alpha$ for each possibility distribution of $\Pi(\Delta,W)$, namely iff:*
$$\forall \pi \in \Pi(\Delta,W), \Pi(\alpha\wedge\beta) > \Pi(\alpha\wedge\neg\beta).$$

In such a case we also write $\alpha \models_{\forall\pi,(\Delta,W)} \beta$, or more simply $\alpha \models_{\forall\Pi} \beta$ where $\alpha$ is viewed as a particular situation to which the knowledge $(\Delta,W)$ is applied in order to deduce plausible conditions. This is indeed a preferential entailment à la Shoham (1988) since it can be checked that $\Pi(\alpha\wedge\beta) > \Pi(\alpha\wedge\neg\beta) \Leftrightarrow \{\omega\models\alpha \mid \pi(\omega) = \Pi(\alpha)>0\} \subseteq \{\omega|\omega\models\beta\}$, i.e., the preferred models of $\alpha$ (which maximize $\pi$) are models of $\beta$. It has been recently established (Dubois and Prade, 1995) that the inferential power of $\models_{\forall\Pi}$ is exactly the one of Kraus, Lehmann and Magidor (1990)' system **P**.

## 2.3. Characterizing $\Pi(\Delta,W)$

In this section, some of noticeable features of the structure of the set $\Pi(\Delta,W)$ are pointed out. For this aim, we associate to each possibility distribution $\pi$ its qualitative counterpart, denoted by $>_\pi$ and called qualitative possibility distribution, defined by $\omega>_\pi\omega'$ iff $\pi(\omega) > \pi(\omega')$, which can be viewed as a well-ordered partition[1] $\{E_1,...,E_n,E_\bot\}$ of $\Omega$ such that:

$$\forall\omega\in E_i, \forall\omega'\in E_j, \pi(\omega)>\pi(\omega') \text{ iff } i<j \text{ (for } i\leq n, j\geq 0).$$

By convention, $E_\bot$ is the set of impossible worlds (i.e., $\forall\omega\in E_\bot, \pi(\omega)=0$). Note that all the possibility distributions $\pi$ of $\Pi(\Delta,W)$ have the same $E_\bot$ (induced by W).

Note that each possibility distribution $\pi$ has exactly one qualitative counterpart $>_\pi$. Two possibility distributions are said to be equivalent if they induce the same partition of $\Omega$, and hence they infer the same set of conclusions by the preferential entailment of Section 2.1. We denote by $Q\Pi(\Delta,W)$ the set of all the qualitative counterparts of the possibility distributions in $\Pi(\Delta,W)$.

The following (easy) proposition shows that, for a qualitative possibility distribution $>_\pi$ of $Q\Pi(\Delta,W)$, splitting any $E_i$ into two layers leads again to a compatible qualitative possibility distribution, i.e.,

**Proposition 1.** *Let $>_\pi = \{E_1, ..., E_i,..., E_n, E_\bot\} \in Q\Pi(\Delta,W)$. Let $>_{\pi'} = \{E_1,...,E'_i,E''_i,...,E_n,E_\bot\}$ obtained from $>_\pi$ by splitting $E_i$ in $E'_i\cup E''_i$. Then $>_{\pi'}$ belongs to*

---

[1] i.e., $\Omega=E_0\cup...\cup E_n\cup E_\bot$, and for $i\neq j$ we have $E_i\cap E_j=\varnothing$, and $\forall i, E_i\cap E_\bot=\varnothing$.

$Q\Pi(\Delta, W)$.

The following definition introduces the notion of qualitative linear possibility distribution:

**Definition 3.** *A qualitative possibility distribution $>_\pi = \{E_1,...,E_n, E_\perp\}$ of $Q\Pi(\Delta, W)$ is said to be linear iff each $E_i \neq E_\perp$ is a singleton (i.e., contains exactly one interpretation).*

**Proposition 2.** *If $(\Delta, W)$ is consistent then there exists at least one linear possibility distribution in $Q\Pi(\Delta, W)$.*

Now, let us introduce the specificity ordering between qualitative possibility distributions (which will be used in Section 2.5. for defining rational closure inference).

**Definition 4.** *Let $>_\pi = \{E_1, ..., E_n, E_\perp\}$ and $>_{\pi'} = \{E'_1, ..., E'_{n'}, E_\perp\}$ two possibility distributions of $Q\Pi(\Delta, W)$. $>_\pi$ is said to be less specific than $>_{\pi'}$ iff*
$$\forall j = 1, max(n,n'), \bigcup_{i=1,j} E'_i \subseteq \bigcup_{i=1,j} E_i$$
*(for $j > min(n,n')$ we use $E_j = \emptyset$ for $n < n'$).*

We denote by $Q\Pi_{max}(\Delta, W)$ the set of the most specific possibility distributions in $Q\Pi(\Delta, W)$. The following proposition shows that the qualitative possibility distributions in $Q\Pi_{max}(\Delta, W)$ are exactly those which are linear in $Q\Pi(\Delta, W)$:

**Proposition 3.** *A possibility distribution $>_\pi$ of $Q\Pi(\Delta, W)$ is linear iff it belongs to $Q\Pi_{max}(\Delta, W)$.*

Let $>_\pi = \{E_1, ..., E_n, E_\perp\}$ and $>_{\pi'} = \{E'_1, ..., E'_m, E_\perp\}$ two possibility distributions of $Q\Pi(\Delta, W)$, then we define the operator Max in the following way:

$$Max(>_\pi, >_{\pi'}) = \{E''_1, ..., E''_{min(n,m)}, E_\perp\}$$

such that $E''_\perp = E_\perp \cup E'_\perp$, $E''_1 = E_1 \cup E'_1$ and

$E''_k = (E_k \cup E'_k) - (\bigcup_{i=1,k-1} E''_i)$ for $i=2, min(n,m)$.

The following proposition shows that the maximum of two qualitative possibility distributions in $Q\Pi(\Delta, W)$ is also in $Q\Pi(\Delta, W)$, i.e.,

**Proposition 4.** *Let $>_\pi$ and $>_{\pi'}$ be two elements of $Q\Pi(\Delta, W)$. Then $Max\{>_\pi, >_{\pi'}\} \in Q\Pi(\Delta, W)$ and $Max\{>_\pi, >_{\pi'}\}$ is less specific than $>_\pi$ and $>_{\pi'}$.*
**Corollary.** *There exists exactly one possibility distribution in $Q\Pi(\Delta, W)$ which is the least specific one, denoted by $>_{\pi spe}$, and defined in the following way:*
$$>_{\pi spe} = Max\{>_{\pi i} / >_{\pi i} \in Q\Pi_{max}(\Delta, W)\}.$$

### 2.4  Two Propositions to Restrict $\Pi(\Delta, W)$

The universal possibilistic consequence relation $\models_{\forall\Pi}$ produces acceptable and safe conclusions, as Lehmann et al.' system P does, but is very cautious. In (Benferhat et al., 1996) it has been proposed to augment its inferential power safely by adding further reasonable constraints to restrict the set of possibility distributions compatible with our beliefs.

Let us first consider the *"irrelevance"* problem. It can be described in the following way: if a formula $\delta$ is a plausible consequence of $\alpha$, and if a formula $\beta$ has "nothing to do" (namely is irrelevant to) with $\alpha$ or $\delta$, then $\delta$ is deduced from $\alpha \wedge \beta$. "nothing to do" is understood here as $\beta$ is a formula composed of propositional symbols which do not appear in the database. This situation is illustrated by the following example:

**Example 1.** *Let us only consider one default rule $\Delta = \{b \rightarrow f\}$. The universal possibilistic consequence relation cannot infer that red birds fly. Indeed, assume that our language contains only three propositional symbols $\{b,f,r\}$ where r is for red, then: $\Omega = \{\omega_0: \neg b \wedge \neg f \wedge \neg r, \omega_1: \neg b \wedge \neg f \wedge r, \omega_2: \neg b \wedge f \wedge \neg r, \omega_3: \neg b \wedge f \wedge r, \omega_4: b \wedge \neg f \wedge \neg r, \omega_5: b \wedge \neg f \wedge r, \omega_6: b \wedge f \wedge \neg r, \omega_7: b \wedge f \wedge r\}$. We can easily check that the three possibility distributions:*
- $\pi_1(\omega_6) = \pi_1(\omega_7) = 1, \pi_1(otherwise) = \alpha < 1$
- $\pi_2(\omega_6) = 1; \pi_2(\omega_5) = \alpha < 1, \pi_2(otherwise) = \beta < \alpha$
- $\pi_3(\omega_6) = 1; \pi_3(\omega_7) = \pi_3(\omega_5) = \alpha < 1, \pi_3(otherwise) = \beta < \alpha$

*are all compatible with $(\Delta, W)$ (since for each $\pi_{i=1,3}$ we have: $\Pi_i(b \wedge f) > \Pi_i(b \wedge \neg f)$). We can easily verify that:*
$$b \wedge r \models_{\pi 1} f, b \wedge r \models_{\pi 2} \neg f \text{ and } b \wedge r \models_{\pi 3} \neg f, b \wedge r \not\models_{\pi 3} f.$$

Clearly, in the previous example the possibility distributions $\pi_2$ and $\pi_3$ are not desirable. It means that another constraint must be added in order to select a subset of $\Pi(\Delta, W)$. For this aim, an interpretation $\omega$ of $\Omega$ is viewed as a pair of conjuncts $\omega = x \wedge y$ where x is an interpretation only constructed from propositional symbols appearing in $\Delta$ or W while y is an interpretation only constructed from propositional symbols which does not appear in $\Delta$ or W, Then we have:

**Definition 5.** *A possibility distribution $\pi$ of $\Pi(W, \Delta)$ is said to cope with irrelevance w.r.t. $(W, \Delta)$ iff for each interpretation $\omega = x \wedge y$ and $\omega' = x' \wedge y'$ we have:*
$$if \ x = x' \ then \ \pi(\omega) = \pi(\omega').$$

It means that if no constraint bears on a symbol s, an interpretation $\omega$ refined by having s true or false, cannot lead to two different levels of possibility. We denote by $\Pi_\mathbb{R}(\Delta, W)$ the set of possibility distributions of $\Pi(\Delta, W)$ which are coping with irrelevance w.r.t. $(\Delta, W)$. The new inference relation, denoted by $\models_{\forall\Pi R}$, is defined as: $(\Delta, W) \models_{\forall\Pi R} \alpha \rightarrow \beta$ iff $\forall \pi \in \Pi_\mathbb{R}(\Delta, W), \Pi(\alpha \wedge \beta) > \Pi(\alpha \wedge \neg \beta)$.

Let $\mathcal{L}_{V(\Delta, W)}$ be the set of all propositional formulas composed of propositional symbols appearing only in $(\Delta, W)$, then we have:

**Proposition 5.** *If $(\Delta, W) \models_{\forall\Pi} \alpha \rightarrow \beta$ and $\delta \notin \mathcal{L}_{V(\Delta, W)}$ then $(\Delta, W) \models_{\forall\Pi R} \alpha \wedge \delta \rightarrow \beta$.*

Using Proposition 5, it is now possible to infer in Example 1 that "red birds fly" using $\models_{\forall\Pi R}$.

A second restriction of $\Pi(\Delta, W)$ is proposed in (Benferhat et al., 1996) in order to recover all the classical entailments obtained from $\{\alpha\} \cup \Delta^* \cup W$, when the observation $\alpha$ is consistent (in a classical sense) with our beliefs, and where $\Delta^*$ is the set of formulas obtained by turning rules in $\Delta$ into strict rules. For example, from the two default rules $\Delta = \{b \rightarrow f, l \rightarrow w\}$, where the second rule reads "generally, animal having legs walk", we would like to deduce that a bird having legs flies. The universal possibilistic consequence relation cannot infer it. This type of conclusions can be obtained by a further restriction of $\Pi(\Delta,W)$:

**Definition 6.** *A possibility distribution $\pi$ of $\Pi_R(\Delta,W)$ is said to be classically consistent with $(W,\Delta)$ iff for each interpretation $\omega$ which is a model of $W \cup \Delta^*$ we have $\pi(\omega) = 1$, and $\pi(\omega) < 1$ otherwise.*

We denote by $\Pi_{RC}(\Delta,W)$ a sub-set of $\Pi_R(\Delta,W)$ of possibility distributions which are consistent w.r.t. $(\Delta,W)$. The inference relation, denoted by $\models_{\forall\Pi RC}$, is defined as: $(\Delta,W) \models_{\forall\Pi RC} \alpha \rightarrow \beta$ iff

$$\forall \pi \in \Pi_{RC}(\Delta,W), \Pi(\alpha \wedge \beta) > \Pi(\alpha \wedge \neg \beta).$$

Then, we have:

**Proposition 6.** *Let $\alpha$ be a formula consistent with $\Delta^* \cup W$. Then:*

$$\{\alpha\} \cup \Delta^* \cup W \vdash \beta \text{ iff } (\Delta,W) \models_{\forall\Pi RC} \alpha \rightarrow \beta.$$

Using Proposition 6, it is now possible to apply transitivity to defaults when no inconsistency in the classical sense takes place. For instance, letting $\alpha=b$ and $\Delta=\{b \rightarrow fo, fo \rightarrow f\}$ where fo is for flying objects, it is now possible to deduce that birds fly.

The inference $\models_{\forall\Pi RC}$ is still cautious in the sense that some expected results cannot be obtained as exemplified in (Benferhat et al., 1996).

### 2.5 Using the Least Specific Distribution and its Limits

There is a radical way to cope with the cautiousness of $\models_{\forall\Pi}$. It is to pick only one possibility distribution among $\Pi(\Delta,W)$: the greatest solution of the set of constraints, which is also said to be the least specific one. The idea is to consider each interpretation as normal as possible, namely to assign to each world $\omega$ the highest possibility level without violating the constraints. The minimum specificity principle which selects the greatest possibility distribution leads to a unique partition of the set of interpretations into different layers $E_1, E_2, ..., E_\perp$ of decreasing possibility as explained in Section 2.3. An algorithm for building this partition from the set of constraints is described in (Benferhat et al., 1992); it is closely related to Pearl (1990)'s Z system for ordering defaults. Then, using the minimum specificity principle leads to defining the following inference relation:

**Definition 7.** *$\beta$ is said to be a MSP-consequence (MSP: for minimum specificity principle) of $\alpha$ w.r.t. to $(\Delta,W)$, denoted by $(\Delta,W) \models_{MSP} \alpha \rightarrow \beta$ iff $\Pi^*(\alpha \wedge \beta) > \Pi^*(\alpha \wedge \neg \beta)$, where $\Pi^*$ is the greatest solution of the constraints associated with $(\Delta,W)$.*

It can be checked that $\pi^*$, associated to $\Pi^*$, belongs to $\Pi_{RC}(\Delta, W)$. The following implications are valid: $(\Delta,W) \models_{\forall\Pi} \alpha \rightarrow \beta \Rightarrow (\Delta,W) \models_{\forall\Pi R} \alpha \rightarrow \beta \Rightarrow (\Delta,W) \models_{\forall\Pi RC} \alpha \rightarrow \beta \Rightarrow (\Delta,W) \models_{MSP} \alpha \rightarrow \beta$.

However the MSP-entailment has still some limitations. A first one, addressed in (Benferhat et al., 1994), is that MSP-entailment is in some cases still too cautious. An (important) case of cautiousness is the so-called "blocking of property inheritance". It corresponds to the case where a class is exceptional for a superclass with respect to some attributes. Then, the least specific possibility distribution does not allow to conclude anything about whether this class is normal or not with respect to other attributes. For example let us expand the penguin example (birds fly, penguins are birds, penguins do not fly) by a fourth default expressing that "birds have legs". As it can be checked, we cannot deduce that a penguin has legs. A natural idea to overcome this cautiousness is to express some kind of independence between the fact of having legs and the fact of flying. This is discussed in the next section.

A second drawback is that the MSP-entailment can produce some conclusions which are not intuitive. Let us consider a variant of "ambiguous" database known as "Nixon diamond" example: "Republicans normally are not pacifists" and "Quakers normally are pacifists". The MSP-entailment can say nothing if Nixon, who is a republican and a quaker, is a pacifist or not. This is intuitively satisfying. Now let us add three further rules (not related to pacifism) to the previous example, which give more information about quakers: "all quakers are Americans", "Americans normally like base-ball" but "all quakers do not like base-ball". This is illustrated by the following figure:

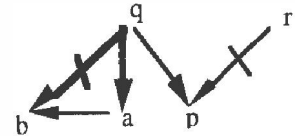

Figure 1

where a "bold" edge represents a strict rule. As explained in (Benferhat et al., 1996), the MSP-entailment will deduce the undesirable result that "republican quakers are pacifists" in the above example. However, it is possible to repair the knowledge base given by an expert, and thus to avoid the deduction of undesirable results. This issue will be discussed in Section 4.

## 3 USING INDEPENDENCE

We are interested in a possibility theory-based encoding of pieces of information of the form "in the context $\alpha$, $\delta$ has no influence on $\beta$" (equivalently, "in the context $\alpha$, $\beta$ is independent of $\delta$ (or irrelevant to $\delta$)"), where $\alpha, \beta, \delta$ are propositional formula; we denote this independence information by $I(\delta, \beta|\alpha)$. The meaning of this statement is that in the context $\alpha$, learning $\delta$ does not change our belief about $\beta$. Otherwise we say that $\delta$ has an influence on $\beta$ (or is relevant to $\beta$) in the context $\alpha$, and we denote it by $R(\delta, \beta|\alpha)$. The next sub-section recalls some properties that have been proposed in the literature for (ir)relevance relations. Next, we propose two related definitions of independence in the possibility theory framework. Finally, we show how the adding of independence information can benefit to default reasoning.

### 3.1 Properties of Independence Relation

The notion of irrelevance is widely used in different areas (Greiner and Subramanian, 1994) and it is not easy to give rational postulates for irrelevance relation since this notion is very dependent on the application that is considered. In this sub-section, we just recall properties proposed by (Subramanian and Genesereth, 1987) since they have considered general postulates for irrelevance relation, and the postulates discussed by Delgrande and Pelletier (1994) since these authors have addressed the problem of irrelevance in the context of default reasoning.

Subramanian and Genesereth (1987) in their attempt to define a notion of irrelevance in problem-solving systems have suggested that irrelevance information must satisfy the following (non)-properties:

- $SG_1$: Non-monotonicity: in general, if in the context $\alpha$, $\beta$ has no influence on $\delta$, then it is not always true that in the context $\alpha \wedge \rho$, $\beta$ still has no influence on $\delta$.
- $SG_2$: Asymmetry: in general, if in the context $\alpha$, $\beta$ has no influence on $\delta$, it does not imply that $\delta$ has also no influence on $\beta$.
- $SG_3$: Intransitivity: in general, independence relation is not transitive.

Delgrande and Pelletier (1994) have addressed the problem of irrelevance in the context of default reasoning. They view the notion of irrelevance as a relation between a property and a conditional assertion. Let $\vdash$ be a nonmonotonic inference relation. Delgrande and Pelletier give some principles for (ir)relevance:

**AUG:**    if $R(\delta, \beta|\alpha)$ then $R(\delta \wedge \rho, \beta|\alpha)$.
**$DeP_1$:** if $I(\delta, \beta|\alpha)$, $\models \alpha \equiv \phi$, $\models \beta \equiv \psi$, $\models \delta \equiv \xi$ then $I(\xi, \psi|\phi)$.
**$DeP_2$:** if $T \vdash \alpha \rightarrow \delta$ then $I(\delta, \beta|\alpha)$.
**$DeP_3$:** if $\models \neg \alpha \vee \neg \delta$ then $I(\delta, \beta|\alpha)$.
**$DeP_4$:** if $R(\delta, \beta|\alpha)$ then $I(\neg \delta, \beta|\alpha)$.
**$DeP_5$:** if $R(\delta, \beta|\alpha)$ then $R(\delta, \neg \beta|\alpha)$.
**$DeP_6$:** if $T \vdash \alpha \rightarrow \beta$, $T \vdash \delta \rightarrow \neg \beta$ and $\alpha \wedge \delta$ is consistent then $R(\delta, \beta|\alpha)$ or $R(\alpha, \neg \beta|\delta)$.

Now, let us explain briefly these properties. First, relevance and irrelevance are assumed to be complementary notions, i.e., $R(\delta, \beta|\alpha)$ iff not($I(\delta, \beta|\alpha)$). The property AUG means that relevance relation is monotonic. $DeP_1$ means that (ir)relevance relation must not be sensitive to the syntax of the premises or the conclusions. $DeP_2$ is similar to the right weakening property described in system P. $DeP_3$ is reasonable if we accept to infer anything from inconsistent premises. $DeP_4$ means that if $\delta$ has an influence on $\beta$ in the context $\alpha$, then $\neg \delta$ is irrelevant to $\beta$ in the context $\alpha$. $DeP_5$ means that irrelevance relation must be insensitive to negation. $DeP_6$ is similar to the cautious monotony of system P.

### 3.2 Possibilistic counterpart of Delgrande and Pelletier's definition of independence

Delgrande and Pelletier define a notion of *relevance* as:

**Definition 8.** $\delta$ *is relevant to* $\alpha \rightarrow \beta$ *(i.e., $R(\delta, \beta|\alpha)$) iff one of the two conditions is satisfied:*
  i) $T \vdash \alpha \rightarrow \beta$ *but* $\alpha \wedge \delta \rightarrow \neg \beta$, *or,*
  ii) $T \vdash \alpha \rightarrow \neg \beta$ *but* $\alpha \wedge \delta \rightarrow \beta$
*where $T$ is the knowledge base, and $\vdash$ is based on some underlying conditional logic.*

A direct way to get a possibilistic counterpart of Delgrande and Pelletier's definition of *independence* is to interpret $T \vdash \alpha \rightarrow \beta$ by $\Pi(\alpha \wedge \beta) > \Pi(\alpha \wedge \neg \beta)) \Leftrightarrow N(\beta|\alpha) > 0$. Then, we get: $\delta$ does not question the acceptance of $\beta$ in the context $\alpha$ iff:

$$(N(\beta|\alpha) = 0 \text{ or } N(\neg\beta|\alpha \wedge \delta) = 0)$$
and   $(N(\neg\beta|\alpha) = 0 \text{ or } N(\beta|\alpha \wedge \delta) = 0)$    (DPInd)

or equivalently, iff:

i)   $N(\beta|\alpha) = N(\neg\beta|\alpha) = 0$, or
ii)  $N(\beta|\alpha \wedge \delta) = N(\neg\beta|\alpha \wedge \delta) = 0$, or
iii) $N(\beta|\alpha) = N(\beta|\alpha \wedge \delta) = 0$, or
iv)  $N(\neg\beta|\alpha) = N(\neg\beta|\alpha \wedge \delta) = 0$.

In other words, if we are in situation of ignorance about $\beta$ either in the context $\alpha$ or $\alpha \wedge \delta$, or if $\beta$ or $\neg \beta$ is totally uncertain both in the context $\alpha$ and $\alpha \wedge \delta$ then $\delta$ has no influence on $\beta$ in the context $\alpha$. This definition in our opinion does not completely match our intuition. For instance, assume that, in the context $\alpha$, we do not know about $\beta$ or $\neg \beta$, and assume that if furthermore learning $\delta$ allows us to deduce $\alpha$. Then intuitively, we like to say that $\delta$ has some (positive) influence on $\beta$. However, using Delgrande and Pelletier's definition we will get that $\delta$ has no influence on $\beta$ according to (i).

There is another reason why this is not the definition of independence that we are looking for. Our first motivation in using the independence information is to increase the inferential power of the nonmonotonic inference relation described in the Section 2. However, if we use (DPInd), then we will still have the blocking inheritance problem To illustrate our saying, consider the penguin example containing the following rules $\{b \rightarrow f, b \rightarrow l, p \Rightarrow b, p \rightarrow \neg f\}$. The use of the MSP-entailment leads to the following partition of $\Omega$:

$E_1 = \{\omega_0: \neg p \wedge \neg b \wedge \neg f \wedge \neg l, \omega_1: \neg p \wedge \neg b \wedge \neg f \wedge l,$
      $\omega_2: \neg p \wedge \neg b \wedge f \wedge \neg l, \omega_3: \neg p \wedge \neg b \wedge f \wedge l, \omega_4: \neg p \wedge b \wedge f \wedge l\}$
$E_2 = \{\omega_5: \neg p \wedge b \wedge \neg f \wedge \neg l, \omega_6: \neg p \wedge b \wedge \neg f \wedge l,$
      $\omega_7: \neg p \wedge b \wedge f \wedge \neg l, \omega_8: p \wedge b \wedge \neg f \wedge l, \omega_9: p \wedge b \wedge \neg f \wedge \neg l\}$
$E_3 = \{\omega_{14}: p \wedge b \wedge f \wedge \neg l, \omega_{15}: p \wedge b \wedge f \wedge l\}.$
$E_\bot = \{\omega_{10}: p \wedge \neg b \wedge \neg f \wedge \neg l, \omega_{11}: p \wedge \neg b \wedge \neg f \wedge l,$
      $\omega_{12}: p \wedge \neg b \wedge f \wedge \neg l, \omega_{13}: p \wedge \neg b \wedge f \wedge l\}.$

It can be easily seen that it is not be possible to deduce that a penguin has legs since the two preferred models (where penguin is true) are $\omega_8$ and $\omega_9$, i.e., one where l is true and one where l is false. And, the previous partition satisfies the irrelevance information using (PDInd) saying that having legs is independent of being a penguin in the context of bird, since:

$$N(l \mid p \wedge b) = N(\neg l \mid p \wedge b) = 0.$$

The problem is that the definition of relevance proposed by Delgrande and Peletier is very strong since we jump from the situation where $\beta$ is accepted to the extreme situation where its contrary $\neg \beta$ is accepted. We shall propose more flexible conditions in the next sub-section.

### 3.3 Possibilistic Independence

Several possible expressions of independence in terms of conditional necessity have been discussed (see Benferhat et al., 1994; Dubois et al., 1994). We only consider a weak independence notion in the following, which is appropriate for default reasoning. Namely $\delta$ is weakly independent of $\beta$ in the context $\alpha$ iff

$$N(\beta|\alpha) > 0 \Rightarrow N(\beta|\alpha \wedge \delta) > 0. \qquad \text{(WInd)}$$

When (WInd) is satisfied we say that $\beta$ is weakly independent of $\delta$ in the context $\alpha$. Note that (WInd) is sensitive to negation, i.e., we cannot change $\delta$ into $\neg\delta$ in (WInd), contrary to probabilistic independence. Similarly, $\beta$ and $\delta$ do not play symmetric roles as is the case in probability theory. The condition (WInd) is also equivalent to

$$\Pi(\beta \wedge \alpha) > \Pi(\neg\beta \wedge \alpha) \Rightarrow \Pi(\beta \wedge \alpha \wedge \delta) > \Pi(\neg\beta \wedge \alpha \wedge \delta).$$

In other words, assuming that the default $\alpha \to \beta$ is independent of the truth of $\delta$ just amounts to supplementing $\alpha \to \beta$ with a more specific default, namely $\alpha \wedge \delta \to \beta$. Two defaults $\alpha \wedge \delta \to \beta$ and $\alpha \wedge \neg\delta \to \beta$ must be added if $\alpha \to \beta$ is claimed to hold regardless of the truth or falsity of $\delta$.

There is one objection against the definition of weak independence. Indeed, it considers that everything is independent of a given formula if the latter is known to be completely uncertain. The following definition of independence (called strong independence) remedies this drawback: $\delta$ does not question the acceptance of $\beta$ in the context $\alpha$ iff

$$N(\beta|\alpha) > 0 \text{ iff } N(\beta|\alpha \wedge \delta) > 0. \qquad \text{(SInd)}$$

It is clear that SInd implies Wind and that SInd implies DPind.[1] Moreover, (SInd) enables us to still guarantee that $N(\beta|\alpha \wedge \delta) = 0$ if $N(\beta|\alpha) = 0$ already holds in the base. Before showing how possibilistic independence can overcome the cautiousness of MSP-entailment, let us see the properties of the weak (strong) independence. The following proposition shows that the three natural non-properties suggested by Subramanian and Genesereth (1987) are satisfied:

**Proposition 7.** *The relation of Weak Independence is non-monotone, asymmetric and intransitive.*
**Counter-examples.** Let p, q, r, s be propositional symbols.
• For non-monotonicity, it is enough to consider the following possibility distribution $\pi$:
$\pi(p \wedge q \wedge s \wedge \neg r) = 1$; $\pi(p \wedge \neg q \wedge s \wedge r) = a<1$; $\pi(\text{otherwise}) = b<a$.
Then $N(q|p) > 0$, $N(q|p \wedge s) > 0$ but $N(q|p \wedge s \wedge r) = 0$.
• For the asymmetry property, consider the following possibility distribution $\pi$:
$\pi(p \wedge \neg q \wedge s) = 1$;

---

[1] In (Dubois et al., 1994), a slightly different definition of independence is discussed. Namely the independence of $\delta$ w.r.t. $\beta$ in the context $\alpha$ amounts to write $N(\beta|\alpha)>0$ *and* $N(\beta|\alpha \wedge \delta)>0$. With this definition, adding to $\Delta$ the supplementary information that $\delta$ is independent from $\beta$ in context $\alpha$, would then include the piece of knowledge $N(\beta|\alpha)>0$, i.e., the rule "if $\alpha$ then plausibly $\beta$", which might be strange, if this piece is not already in $\Delta$ (since it is not clear that people stating independence information also intend to implicitly state other default information). If $N(\beta|\alpha)>0$ is already in $\Delta$, there is no difference with the weak or the strong independence defined above.

$\pi(\text{otherwise}) = a<1$.
we have $N(q|p)=0$ (hence s has no influence on q), while $N(s|p)>0$ but $N(s|p \wedge q)=0$.
• The weak independence relation is not transitive, indeed:
$\pi(p \wedge \neg r \wedge q \wedge s) = \pi(p \wedge \neg r \wedge q \wedge \neg s) = 1$,
$\pi(p \wedge r \wedge \neg q \wedge s) = b<1$,
$\pi(\text{otherwise}) = c<b$.
We can check that:
- $N(q|p) > 0$ and $N(q|p \wedge s) > 0$; i.e., q is weakly independent of s in the context p,
- $N(s|p) = 0$; i.e., s is weakly independent of r in the context p, but:
- $N(q|p) > 0$ and $N(q|p \wedge r) = 0$.

**Corollary.** *The relation of Strong Independence is non-monotone, asymmetric and intransitive.*
**Proof.** It is enough to consider the counter-examples above.

Defining Weak Relevance $R(\delta, \beta|\alpha)$ as a negation of the definition of weak independence, we obtain:

"$\delta$ is weakly relevant to $\beta$ in the context $\alpha$ iff
$N(\beta|\alpha) > 0$ and $N(\beta|\alpha \wedge \delta)=0$".

Similarly, we define strong relevance as: "$\delta$ is strongly relevant to $\beta$ in the context $\alpha$ iff $(N(\beta|\alpha)>0$ and $N(\beta|\alpha \wedge \delta)=0)$ or $(N(\beta|\alpha)=0$ and $N(\beta|\alpha \wedge \delta)>0)$".

Then we can show that the property AUG proposed in (Delgrande and Pelletier, 1994) is not satisfied by any of the two proposed definitions of independence. Indeed let us consider the following example:

$\pi(p \wedge r \wedge q \wedge \neg s) = 1$;   $\pi(p \wedge \neg r \wedge \neg q \wedge s) = b<1$,
$\pi(p \wedge r \wedge q \wedge s) = c<b$,   $\pi(\text{otherwise}) = d<c$.

We can see that $N(q|p)>0$, $N(q|p \wedge s)=0$ (i.e., s is weakly (hence strongly) relevant to q in the context p) but $N(p|q \wedge s \wedge r)>0$. Moreover, we have:

**Proposition 8.** *The relation of weak independence satisfies $DeP_{1,2,4,6}$ and fails to satisfy $DeP_{3,5}$.*
**Proof.**
• $DeP_1$ is obvious since possibilistic entailment is independent of the syntax of the formulas.
• For $DeP_2$, let us assume that $N(\delta | \alpha) > 0$. We have two cases:
- either $N(\beta | \alpha) = 0$ then obviously $\beta$ is independent of $\delta$ in the context $\alpha$,
-or $N(\beta|\alpha)>0$, then with $N(\delta|\alpha)>0$ we deduce that $N(\beta \wedge \delta|\alpha)>0$ which is equivalent to $\Pi(\alpha \wedge \beta \wedge \delta) > \Pi(\alpha \wedge (\neg\beta \vee \neg\delta))$ which implies $\Pi(\alpha \wedge \beta \wedge \delta) > \Pi(\alpha \wedge \neg\beta)$, hence $\Pi(\alpha \wedge \beta \wedge \delta) > \Pi(\alpha \wedge \neg\beta \wedge \delta)$ and thus $N(\beta|\alpha \wedge \delta)>0$. Note that $DeP_2$ is similar to the cautious monotony (Kraus et al., 1990).
• $DeP_3$ does not hold. Consider the following possibility distribution:
$\pi(\alpha \wedge \beta \wedge \neg\delta) = 1$ and $\pi(\alpha \wedge \beta \wedge \delta) = 0$
$\pi(\text{otherwise}) = \alpha<1$ and $\pi(\alpha \wedge \neg\beta \wedge \delta) = 0$
then we have:
$N(\neg\alpha \vee \neg\delta) = 1$, $N(\beta|\alpha)>0$ and $N(\beta|\alpha \wedge \delta) = 0$.
• Proof of $DeP_4$: From $R(\delta, \beta | \alpha)$ we have:
$N(\beta|\alpha) > 0 \Leftrightarrow \Pi(\alpha \wedge \beta) > \Pi(\alpha \wedge \neg\beta)$
$\Leftrightarrow \max(\Pi(\alpha \wedge \beta \wedge \delta), \Pi(\alpha \wedge \beta \wedge \neg\delta))$
$> \max(\Pi(\alpha \wedge \neg\beta \wedge \delta), \Pi(\alpha \wedge \neg\beta \wedge \neg\delta))$ (a)
Moreover: $N(\beta|\alpha \wedge \delta) = 0 \Leftrightarrow \Pi(\alpha \wedge \neg\beta \wedge \delta) \geq \Pi(\alpha \wedge \beta \wedge \delta)$ (b)
(a), (b) $\Rightarrow \Pi(\alpha \wedge \beta \wedge \neg\delta) > \max(\Pi(\alpha \wedge \neg\beta \wedge \delta),$
$\Pi(\alpha \wedge \neg\beta \wedge \neg\delta), \Pi(\alpha \wedge \beta \wedge \delta))$
$\Rightarrow \Pi(\alpha \wedge \beta \wedge \neg\delta) > \Pi(\alpha \wedge \neg\beta \wedge \neg\delta)$
$\Rightarrow N(\beta|\alpha \wedge \neg\delta) > 0$       and hence the thesis.

- $DeP_5$ obviously does not hold since we cannot have $N(\beta|\alpha)>0$ and $N(\neg\beta|\alpha) > 0$. However the following postulates trivially holds:

   if $R(\delta,\beta | \alpha)$ then $I(\delta,\neg\beta|\alpha)$

since from $R(\delta,\beta | \alpha)$ we have $N(\beta|\alpha) > 0$, then $N(\neg\beta|\alpha) = 0$ hence $I(\delta,\neg\beta|\alpha)$.
- $DeP_6$ holds. Indeed, assume that we have $I(\delta,\beta|\alpha)$ and $I(\alpha,\neg\beta|\delta)$. Then with the assumptions that $N(\beta|\alpha) > 0$ and $N(\neg\beta|\delta) > 0$, we deduce that $N(\beta|\alpha\wedge\delta) > 0$ and $N(\neg\beta|\delta\wedge\alpha) > 0$, which is impossible.

**Proposition 9.** *The relation of strong independence satisfies $DeP_{1,2,4,6}$ and fails to satisfy $DeP_{3,5}$.*
**Proof.**
- The proof of $DeP_1$ and counter-examples of $DeP_{3,5}$ are exactly the same as the ones of Proposition 4.
- For $DeP_2$ since the weak independence satisfies this postulate, then it is enough to show that if $N(\beta|\alpha)=0$ then $N(\beta|\alpha\wedge\delta)=0$. Assume that $N(\beta|\alpha\wedge\delta)>0$ then we have $\Pi(\beta\wedge\alpha\wedge\delta)>\Pi(\neg\beta\wedge\alpha\wedge\delta)$. $N(\delta|\alpha)>0$ implies that $\Pi(\delta\wedge\alpha)>\Pi(\neg\delta\wedge\alpha)$, which implies that $\Pi(\beta\wedge\alpha\wedge\delta)>\Pi(\neg\beta\wedge\alpha\wedge\neg\delta)$, Therefore we have $N(\beta|\alpha)>0$, and this is impossible.
- For $DeP_4$, using Proposition 4, it is enough to consider the second case of $R(\delta,\beta|\alpha)$, i.e., $N(\beta|\alpha)=0$ and $N(\beta|\alpha\wedge\delta)>0$. We can show that the previous constraints implies that $R(\neg\delta,\beta|\alpha)$. Indeed, $N(\beta|\alpha\wedge\delta)>0$ and $N(\beta|\alpha\wedge\neg\delta)>0$ imply trivially $N(\beta|\alpha)>0$ which is impossible.
- The proof of $DeP_6$ is obvious using Prop. 4 since when $N(\beta|\alpha)>0$ then weak and strong independence are equivalent.

### 3.4 Application to Plausible Reasoning

Let us first go back to the two situations recalled in Section 2.4 where some intuitive ideas of independence were translated in new constraints added to the knowledge base. Such a treatment is in agreement with the possibilistic view of independence we just introduced. Indeed, in the "irrelevance problem", constraints of the form $\Pi(\alpha\wedge r)=\Pi(\alpha\wedge\neg r)$ were added to $\Delta$ each time r is a partial interpretation built with literals not appearing in $(\Delta,W)$. Such a constraint is equivalent to $N(r|\alpha)=N(\neg r|\alpha)=0$. Then if $N(\beta|\alpha)>0$ encodes a default in $\Delta$, $N(\beta|\alpha)>0$ and $N(\neg r|\alpha)=0$ entail $N(\beta|\alpha\wedge r)>0$ (this is rational monotony, see Benferhat et al., 1992). Thus the added constraints are indeed expressing independence. They are a bit stronger since in the absence of information there is no reason to somewhat relax the equality $\Pi(\alpha\wedge r)=\Pi(\alpha\wedge\neg r)$ by introducing intermediary layers in the partition of $\Delta$ in order to have $\Pi(\alpha\wedge r)>\Pi(\alpha\wedge\neg r)$ for some $\alpha$ and r, in a compatible way with the independence constraints.

In Section 2.4 it was also proposed to enforce constraints of the form $1=\pi(\omega)>\pi(\omega')$ as soon as $\omega\models\Delta^*\cup W$ and $\omega'\not\models\Delta^*\cup W$, where $\omega\models\Delta^*\cup W$ means that $\omega$ is an interpretation which corresponds to a normal situation with respect to all the default rules in $\Delta$. Thus, if $\Pi(\beta\wedge\alpha)>\Pi(\neg\beta\wedge\alpha)$ holds for expressing a default rule, we have still $\Pi(\beta\wedge\alpha\wedge\varphi)>\Pi(\neg\beta\wedge\alpha\wedge\varphi)$ if $\varphi$ is a partial interpretation consistent with $\Delta$ (i.e., $\beta\wedge\alpha\wedge\varphi\models\Delta^*\cup W$) since $\Pi(\beta\wedge\alpha)=\max(\Pi(\beta\wedge\alpha\wedge\varphi),\Pi(\beta\wedge\alpha\wedge\neg\varphi))$. Thus we are expressing independence with respect to situations which are normal with respect to the default rules in the base.

We now present a general procedure for handling default rules and independence together. Let $\mathcal{I} = \{I(\delta_i,\beta_i|\alpha_i) / i=1,m\}$ be a set of weak independencies (for a lack of space, strong independencies are not considered here). A possibility distribution $\pi$ of $\Pi(\Delta,W)$ is said to satisfy the set $\mathcal{I}$ iff for each independence information $I(\delta_i,\beta_i|\alpha_i)$ of $\mathcal{I}$ we have: if $\Pi(\beta_i\wedge\alpha_i)>\Pi(\neg\beta_i\wedge\alpha_i)$ then $\Pi(\beta_i\wedge\alpha_i\wedge\delta_i)>\Pi(\neg\beta_i\wedge\alpha_i\wedge\delta_i)$. The following algorithm computes the partition $E_1, ..., E_i, E_\perp$ of $\Omega$ obtained by applying the minimum specificity principle to the set of possibility distributions of $\Pi(\Delta, W)$ which satisfy $\mathcal{I}$. We denote by:
$\mathcal{C}_\Delta=\{Max(\omega|\omega\models\alpha_i\wedge\beta_i) >_\pi Max(\omega|\omega\models\alpha_i\wedge\neg\beta_i) | \alpha_i\rightarrow\beta_i\in\Delta\}$ the set of constraints on models induced by $\Delta$, by $D(\mathcal{C}_\Delta)$ the set of interpretations which do not appear in the right side of any constraint in $\mathcal{C}_\Delta$. The algorithm is:

a. $i:=0$;
b. Let $E_\perp=\{\omega|\exists\alpha_j \Rightarrow \beta_j\in W \text{ s.t. } \omega\models \alpha_j\wedge\neg\beta_j\}$; $\Omega:=\Omega-E_\perp$.
c. While $\Omega\neq\emptyset$ do c.1.-c.5
   c.1. $i := i+1$; $E_i:=\{\omega / \omega\in\Omega \text{ and } \omega\notin D(\mathcal{C}_>)\}$; $A:=E_i$;
   c.2. Remove from $\mathcal{I}$ each $I(\delta_i, \beta_i|\alpha_i)$ s.t., $\exists\omega\in E_i$ and
        $\omega\models\neg\beta_i\wedge\alpha_i$
   c.3. For each $I(\delta_i, \beta_i|\alpha_i)$ of $\mathcal{I}$ s.t., $\exists\omega\in E_i$ and $\omega\models\beta_i\wedge\alpha_i$
      c.3.1. $\mathcal{C}_\Delta = \mathcal{C}_\Delta \cup$
         $\{Max(\omega|\omega\models\alpha_i\wedge\beta_i\wedge\delta_i) >_\pi Max(\omega|\omega\models\alpha_i\wedge\neg\beta_i\wedge\delta_i)\}$
      c.3.2. $\mathcal{I} = \mathcal{I} - \{I(\delta_i, \beta_i|\alpha_i)\}$
   c.4. If $E_i = \emptyset$ Then Stop (inconsistent beliefs)
   c.5. Remove from $\mathcal{C}_\Delta$ any constraint containing at least
        one interpretation of $E_i$
   c.6. $\Omega := \Omega - E_i$
d. Return $E_1, ..., E_i, E_\perp$.

At each level i, the algorithm tries to put as much as possible interpretations in $E_i$. For this aim, it puts in $E_i$ all interpretations which do not appear in the right side any constraint in $\mathcal{C}_\Delta$ (step c.1.).

Let us go back to the penguin example containing the following rules $\{b\rightarrow f, b\rightarrow l, p\Rightarrow b, p\rightarrow\neg f\}$ and given in the previous section. Let us assume that we have one independence information $I(p,l|b)$ saying that in the context b, the property l does not depend on p. The use of the algorithm described above leads to the following partition of $\Omega$:

$E_1 = \{\omega_0: \neg p\wedge\neg b\wedge\neg f\wedge\neg l, \omega_1: \neg p \wedge \neg b \wedge \neg f\wedge l,$
        $\omega_2: \neg p\wedge\neg b\wedge f\wedge\neg l, \omega_3: \neg p\wedge\neg b\wedge f\wedge l, \omega_4: \neg p\wedge b\wedge f\wedge l\}$
$E_2 = \{\omega_5: \neg p\wedge b\wedge\neg f\wedge\neg l, \omega_6: \neg p\wedge b\wedge\neg f\wedge l,$
        $\omega_7: \neg p\wedge b\wedge f\wedge\neg l, \omega_8: p\wedge b\wedge\neg f\wedge l\}$
$E_3 = \{\omega_{14}: p\wedge b\wedge f\wedge\neg l, \omega_{15}: p\wedge b\wedge f\wedge l, \omega_9: p\wedge b\wedge\neg f\wedge\neg l\}.$
$E_\perp = \{\omega_{10}: p\wedge\neg b\wedge\neg f\wedge\neg l, \omega_{11}: p\wedge\neg b\wedge\neg f\wedge l,$
        $\omega_{12}: p\wedge\neg b\wedge f\wedge\neg l, \omega_{13}: p\wedge\neg b\wedge f\wedge l\}.$

It is now possible to conclude that penguins have legs.

## 4 CONCLUDING DISCUSSION

Starting with the idea that System P is providing a minimal basis for reasoning with default rules and taking advantage of its semantics in terms of a family of possibility distributions, (i.e., in terms of a family of complete rankings of the interpretations), we have shown how to increase its inferential power by adding independence constraints on the possibility distributions. This enables the machinery to unblock intuitive results.

Beside default rules and independence information, a third kind of knowledge has to be added in the knowledge base in the general case. This latter kind of information corresponds to the statement of the form "in the context $\alpha$, do not infer $\beta$" which is encoded in possibility theory by $\Pi(\alpha \wedge \neg \beta) \geq \Pi(\alpha \wedge \beta) \Leftrightarrow N(\beta|\alpha)=0$, and indeed corresponds to another type of constraints on the possibility distributions. Handling such information can help an expert to construct his knowledge base

Indeed, often, the knowledge base provided by an expert is incomplete. Then, the inference machinery has to complete this knowledge, and may infer some conclusions by default which are not desirable for the expert. For instance, assume that an expert only gives one piece of information $\Delta=\{b \rightarrow f\}$. This database is incomplete since it does not tell if birds which live in Antarctica fly or not. MSP-entailment will answer "yes" to this question, since it makes some "closed-world" assumption which enables the machinery to infer such a result. Assume that this result does not please the expert who finds it debatable. The problem is then to find, which rule must be added to the knowledge base such that it will no longer be possible to infer this result. At the semantical level, it is enough to add the constraint:

$$\Pi(b \wedge a \wedge f) \leq \Pi(b \wedge a \wedge \neg f)$$

which means that we ignore if birds living in Antarctica fly or not. The application of minimum specificity principle to the set of constraints $\{\Pi(b \wedge a \wedge f) \leq \Pi(b \wedge a \wedge \neg f), \Pi(b \wedge f) > \Pi(b \wedge \neg f)\}$ leads to the following partition of $\Omega$:

$E_1 = \{\omega_0: \neg b \wedge a \wedge f, \omega_1: \neg b \wedge \neg a \wedge f, \omega_2: \neg b \wedge a \wedge \neg f, \omega_3: \neg b \wedge \neg a \wedge \neg f, \omega_4: b \wedge \neg a \wedge f\}$

$E_2 = \{\omega_5: b \wedge a \wedge \neg f, \omega_6: b \wedge a \wedge f, \omega_7: b \wedge \neg a \wedge \neg f\}$.

Note that this possibility distribution $\pi$ belongs to $\overline{\Pi}(\Delta,W)$, but $\pi$ is neither consistent nor relevant to $\Delta$. Then, it is easy to check that birds fly, but we ignore if birds living in antarctic fly or not.

More generally, it can be shown that any MSP-consequence which is not an universal consequence can be retracted, namely it can be shown (Benferhat et al., 1996):

**Proposition 10** *Let $(\Delta, W) \vDash_{MSP} \phi \rightarrow \psi$. Then if $(\Delta,W) \nvDash_{\forall \Pi} \phi \rightarrow \psi$ then $\phi \rightarrow \psi$ is retractable, otherwise it is not.*

The previous proposition is very important to see how a defeasible reasoning system works. First, an expert gives a set of incomplete knowledge base $(\Delta, W)$. Then we compute a superset of $(\Delta,W)$ given by the possibilistic universal consequence relation. We denote this superset by $(\Delta,W)^P$ (P for system P of Kraus et al. (1990)). The set $(\Delta,W)^P$ only contains conclusions that can be safely inferred from $\Delta$. The conclusions of $(\Delta,W)^P$ are neither debatable nor retractable.

However, there are several ways to enlarge $(\Delta,W)^P$ by choosing one possibility distribution compatible with $(\Delta,W)$. One way to choose one element of $\overline{\Pi}(\Delta,W)$ is to apply the minimum specificity principle. The added results to $(\Delta,W)^P$ by MSP-entailment are all by defaults. Nevertheless, if an expert considers that a conclusion, say $\phi \rightarrow \psi$, is not desirable then it is still possible to correct and repair the initial knowledge base. Repairing here means adding a new information represented by the constraint $\Pi(\phi \wedge \psi) \geq \Pi(\phi \wedge \neg \psi)$ and then to apply the minimum specificity principle to the set of constraints thus augmented.

Let us now go back to our example given in Section 2.4. If we do not want to have that "quakers which are republicans are pacifists", then we impose the constraint:

$$\Pi(q \wedge r \wedge p) = \Pi(q \wedge r \wedge \neg p),$$

Applying the minimum specificity leads to a partition of $\Omega$ where we ignore if "republican which are quakers" are pacifists or not (see (Benferhat et al., 1996) for details).

Thus, undesirable conclusions seem only be due to missing pieces of knowledge that the system cannot guess on its own. Besides what has been discussed remains at the semantical level and syntactic counterparts of the presented procedures have still to be developed.